\documentclass{article} % For LaTeX2e
\usepackage{iclr2024_conference,times}

\usepackage{amsmath,amsfonts,bm}

\def\eqref#1{equation~\ref{#1}}
\def\Eqref#1{Equation~\ref{#1}}
\def\1{\bm{1}}

\def\ve{{\bm{e}}}
\def\vf{{\bm{f}}}

\def\vt{{\bm{t}}}
\def\vu{{\bm{u}}}

\def\vz{{\bm{z}}}

\def\mP{{\bm{P}}}
\def\mQ{{\bm{Q}}}

\def\mU{{\bm{U}}}

\def\mW{{\bm{W}}}

\DeclareMathAlphabet{\mathsfit}{\encodingdefault}{\sfdefault}{m}{sl}
\SetMathAlphabet{\mathsfit}{bold}{\encodingdefault}{\sfdefault}{bx}{n}

\def\emP{{P}}
\def\emQ{{Q}}

\usepackage{hyperref}
\usepackage{url}
\usepackage{graphicx}
\usepackage{cleveref}

\usepackage{makecell}
\usepackage{mathtools}
\usepackage{subcaption}
\usepackage{amsmath}
\usepackage{amsthm}

\theoremstyle{plain}
\newtheorem{theorem}{Theorem}[section]

\newtheorem{hyp}[theorem]{Hypothesis}

\theoremstyle{remark}

\theoremstyle{definition}

\makeatletter
\newcommand{\myfnsymbol}[1]{%
  \expandafter\@myfnsymbol\csname c@#1\endcsname
}
\newcommand{\@myfnsymbol}[1]{%
  \ifcase #1
  \or a% 1
  \or b% 2
  \or c% 3
  \or \TextOrMath{\textdagger}{\dagger}% 4
  \or \TextOrMath{\textasteriskcentered}{*}% 5
  \fi
}
\newcommand{\affiliationCMTC}{\@myfnsymbol{1}}
\newcommand{\affiliationUMD}{\@myfnsymbol{2}}

\newcommand{\correspondingA}{\@myfnsymbol{4}}
\newcommand{\equalcontributor}{\@myfnsymbol{5}}
\makeatother
\title{Grokking Modular Polynomials}

\author{Darshil Doshi \textsuperscript{\normalfont{\affiliationCMTC, \affiliationUMD, \correspondingA}} \\
\And
Tianyu He \textsuperscript{\normalfont{\affiliationCMTC, \affiliationUMD}} \\
\And
Aritra Das \textsuperscript{\normalfont{\affiliationUMD}} \\
\And
Andrey Gromov \textsuperscript{\normalfont{\affiliationCMTC, \affiliationUMD}} \hspace{1em}\\
}

\iclrfinalcopy 
\begin{document}

% Thanks notes for title uses \myfnsymbol
\renewcommand{\thefootnote}{\myfnsymbol{footnote}}
\maketitle
% Layout the \thanks notes in the order you want
\footnotetext[1]{Condensed Matter Theory Center, University of Maryland, College Park}%
\footnotetext[2]{Department of Physics, University of Maryland, College Park}%
\footnotetext[4]{Corresponding author}%

\setcounter{footnote}{0}% Restart footnote counter
% Footnotes for rest of document uses \fnsymbol (or whatever you choose)
\renewcommand{\thefootnote}{\arabic{footnote}}

\vspace{-2.5em}
\hspace{5em}
\texttt{\{ddoshi, tianyuh, aritrad, andrey\}@umd.edu}
\vspace{2em}

\maketitle

% \maketitle

\begin{abstract}
Neural networks readily learn a subset of the modular arithmetic tasks, while failing to generalize on the rest. This limitation remains unmoved by the choice of architecture and training strategies. On the other hand, an analytical solution for the weights of Multi-layer Perceptron (MLP) networks that generalize on the modular addition task is known in the literature.
In this work, we (i) extend the class of analytical solutions to include modular multiplication as well as modular addition with many terms. Additionally, we show that real networks trained on these datasets learn similar solutions upon generalization (grokking). (ii) We combine these ``expert" solutions to construct networks that generalize on arbitrary modular polynomials. (iii) We hypothesize a classification of modular polynomials into learnable and non-learnable via neural networks training; and provide experimental evidence supporting our claims.

\end{abstract}

\section{Introduction}

Modular arithmetic finds applications in many important fields including cryptography \citep{regev2009lattices, wenger2023salsa}, computer algebra \citep{fortin2020high}, number theory, error detection in serial numbers and music \citep{askew2018modular}. Recently, modular arithmetic has peaked the interest of the machine learning community due to the phenomenon of \emph{Grokking} \citep{power2022grokking} -- delayed and sudden generalization that occurs long after memorization. Since it was first observed \citep{power2022grokking}, many works have attempted to understand \emph{grokking dynamics} \citep{liu2022towards, liu2023omnigrok, thilak2022slingshot, notsawo2023predicting, kumar2023grokking, lyu2023dichotomy, davies2022unifying, minegishi2023grokking}. Another line of work attempts to unravel the grokking behaviors using solvable models and mechanistic interpretability \citep{nanda2023progress, gromov2022grokking, zhong2023clock, barak2022hidden, merrill2023twocircuits, doshi2023grok, varma2023explaining, rubin2023droplets}. It was noted in \citet{power2022grokking} that neural networks readily grok a subset of modular arithmetic polynomials while failing to generalize on the rest. This limitation has remained unwavering to variations in architectures and training strategies. 

Notably, \citet{gromov2022grokking} presented an ``analytical solution" for the weights of a 2-layer MLP network that has a $100\%$ accuracy on modular addition dataset ($n_1 + n_2 \, \mathrm{mod} \, p$). They showed that real networks trained on modular addition data finds similar solutions upon grokking. In this work, we extend the class of analytical solutions to to include modular multiplication (e.g. $n_1 n_2 \, \mathrm{mod} \, p$) and modular addition with many terms (e.g. $n_1 + n_2 + \cdots n_S \, \mathrm{mod} \, p$).\footnote{The solution for modular multiplication can be also readily extended to many terms. We present the solution for two terms just for simplicity.} Indeed, we show that training real networks leads to similar network weights upon grokking for both of these tasks. Using these analytical solutions as ``experts", we construct networks that offers generalization on arbitrary modular polynomials, \emph{including the ones that are deemed un-learnable in literature}. This formulation opens up the possibility of training Mixture-of-Experts \citep{jordan1993hierarchical, shazeer2017outrageously, lepikhin2021gshard, fedus2022switch} models that can be trained to learn arbitrary modular arithmetic tasks. Based on our analysis, we hypothesize a classification of modular polynomials into learnable and non-learnable by 2-layer MLPs (and possibly more general architectures such as Transformers and deeper MLPs).

\section{Modular Addition with Many Terms}

Consider the modular addition task with many terms with arbitrary coefficients: $\left( c_1 n_1 + c_2 n_2 + \cdots + c_S n_S \right) \, \mathrm{mod} \, p \,;$
where $c_s \in \mathbb{Z}_p \backslash \{0\}$ are the nonzero coefficients of the modular variables $n_s \in \mathbb{Z}_p$. Note that this is a generalization of the modular addition tasks generally considered in literature: $(n_1+n_2) \, \mathrm{mod} \, p$ \citep{power2022grokking, gromov2022grokking}. 
We consider a 2-layer MLP (of sufficient width) for this task.
\begin{align}\label{eq:net_multisum}
    \vf_{addS} (\ve_{n_1} \oplus \cdots \oplus \ve_{n_S}) = \mW \phi\left( \mU (\ve_{n_1} \oplus \cdots \oplus \ve_{n_S}) \right) = \mW \phi\left( \mU^{(1)} \, \ve_{n_1} + \cdots + \mU^{(S)} \, \ve_{n_S} \right)\,,
\end{align}

where $\ve_{n_1}, \dots, \ve_{n_S} \in \mathbb{R}^p$ are $\texttt{one\_hot}$ encoded numbers $n_1, \dots, n_S$. ``$\oplus$'' denotes concatenation of vectors $(\ve_{n_1} \oplus \cdots \oplus \ve_{n_S} \in \mathbb R^{Sp})$. $\mU \in \mathbb R^{N \times Sp}$ and $\mW \in \mathbb R^{p\times N}$ are the first and second layer weight matrices, respectively. $\phi(x) = x^S$ is the element-wise activation function. $\mU$ is decomposed into $S$ blocks of $N\times P$: $\mU = \mU^{(1)} \oplus \cdots \oplus \mU^{(S)}$. $\mU^{(1)}, \dots, \mU^{(S)}$ serve as embedding matrices for $n_1, \dots, n_S$. $\vf (\ve_{n_1} \oplus \cdots \oplus \ve_{n_S}) \in \mathbb R^p$ is the network-output on one example datapoint $(n_1, \dots, n_S)$. The targets are $\texttt{one\_hot}$ encoded answers $\ve_{(c_1n_1 + \cdots + c_Sn_S)\,\mathrm{mod}\,p}$.

\subsection{Analytical solution}

For a sufficiently wide network, it is possible to write a solution for the weights of the network that generalizes with $100\%$ accuracy.
\begin{align}\label{eq:solution_multisum}
\begin{aligned}
    &U^{(s)}_{ki} = A \, \cos{\left[ \frac{2\pi}{p}\sigma(k) c_s i + \psi^{(s)}_k \right]}&
    &\left(\forall \, s \in [0,S] \right)&\\
    &W_{qk} = A \, \cos{\left[ -\frac{2\pi}{p}\sigma(k) q - \sum_{s=1}^S\psi^{(s)}_k \right]}\,,&
    % & \xi_k = \sum_{t=1}^S\psi^{(t)}_k \,,
\end{aligned}
\end{align}
where $\sigma(k)$ denotes a random permutation of $k$ in $S_N$ -- reflecting the permutation symmetry of the hidden neurons. The phases $\psi_k^{(s)}$ uniformly i.i.d. sampled between $(-\pi, \pi]$. $A$ is the normalization factor to ensure correct magnitude of the output logits, given by $A={\left(2^S/(N \cdot S!)\right)}^{1/(S+1)}$. 
Note that this solution is a generalization of that presented in \citet{gromov2022grokking}.

The rows (columns) of $\mU^{(1)}, \dots, \mU^{(S)}, \mW$ are periodic with frequencies $\frac{2\pi}{p}\sigma(k)c_1, \cdots, \frac{2\pi}{p}\sigma(k)c_S, -\frac{2\pi}{p}\sigma(k)$. We can use Inverse Participation Ratio (IPR) \citep{gromov2022grokking, doshi2023grok} to quantify the similarity between the analytical solution and real trained networks. We denote discrete Fourier transforms of the row- (column-) vectors $\mU^{(s)}_{k\cdot}, \mW_{\cdot k}$ by $\mathcal{F}\left(\mU^{(s)_{k\cdot}}\right), \mathcal{F}(\mW_{\cdot k})$.\footnote{In general IPR for a vector $\vu$ is defined as $(\lVert \vu \rVert_{2r} / \lVert \vu \rVert_{2} )^{2r}$. We set $r=2$, which is a common choice.}
\begin{align}\label{eq:IPR}
    &\text{IPR}\left(\mU^{(s)}_{k\cdot}\right) \coloneqq \left(\frac{\left\lVert \mathcal{F}\left(\mU^{(s)}_{k\cdot}\right) \right\rVert_4}{\left\lVert \mathcal{F}\left(\mU^{(s)}_{k\cdot}\right) \right\rVert_2} \right)^4 \,;&
    &\text{IPR}\left(\mW_{\cdot k}\right) \coloneqq \left(\frac{\left\lVert \mathcal{F} \left(\mW_{\cdot k}\right) \right\rVert_4}{\left\lVert \mathcal{F}\left(\mW_{\cdot k}\right) \right\rVert_2} \right)^4 \,;&
\end{align}
where $\lVert \cdot \rVert_P$ denotes the $L^P$-norm of the vector. Subscripts ``$k\cdot$" and ``$\cdot k$" denote $k^{th}$ row and column vectors respectively. We also define the per-neuron IPR as
\begin{align}
    \text{IPR}_k \coloneqq \frac{1}{S+1} \left( \text{IPR}\left( \mU^{(1)}_{k\cdot} \right) + \cdots + \text{IPR}\left( \mU^{(S)}_{k\cdot} \right) + \text{IPR}\left(\mW_{\cdot k}\right) \right) \,,
\end{align}
and the average IPR of the network by averaging it over all neurons as $\overline{IPR} \coloneqq \mathbb E_k \left[ \text{IPR}_k \right]$.

\begin{figure}[!h]
    \centering
    \includegraphics[width=\columnwidth]{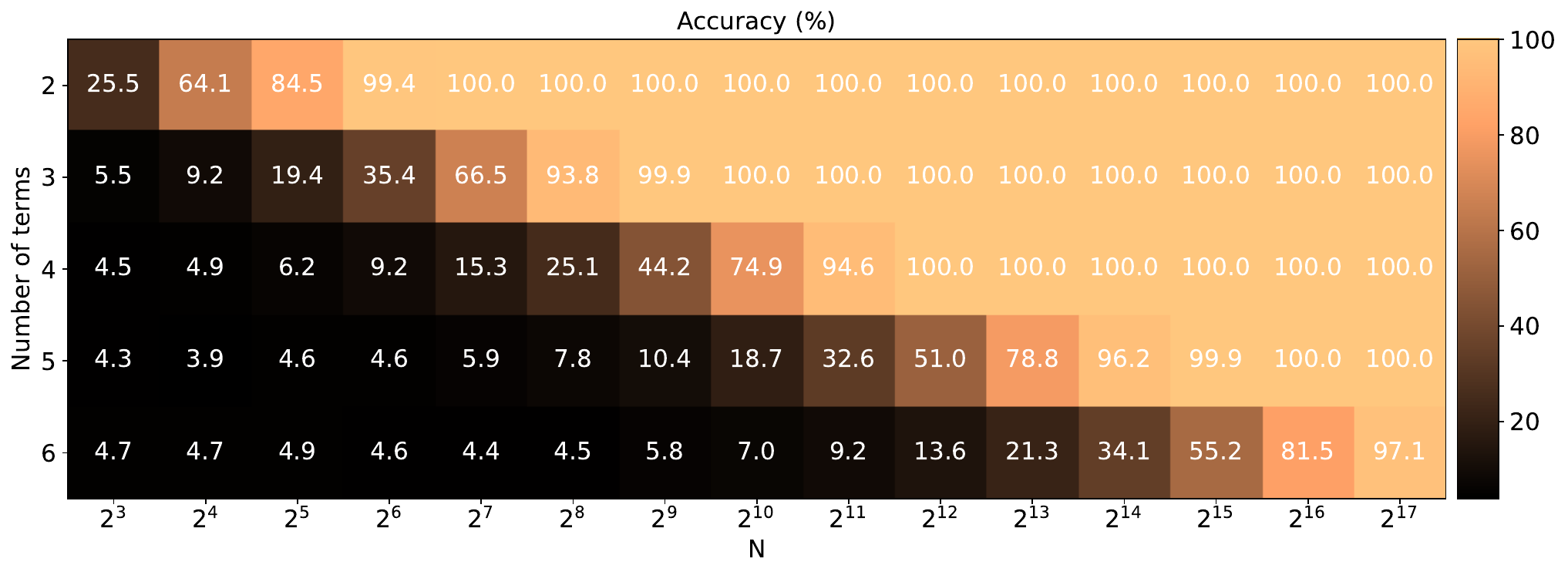}
    \caption{Modular addition with many terms -- analytical solution applied to real 2-layer MLP networks ($p=23$). The solution \eqref{eq:solution_multisum} works for sufficiently wide networks. Note that the x-axis is scaled logarithmically; which suggests an exponential increase in the required width upon adding more terms (as expected). The accuracies shown are calculated over a randomly chosen subset of the entire dataset, consisting of 10k examples. The results shown are the best out of 10 random seeds.}
    \label{fig:multisum_analytical}
\end{figure}

In \Cref{fig:multisum_analytical}, we show that the analytical solution \eqref{eq:solution_multisum} indeed has $100\%$ for sufficient network-width. We observe an exponential increase in the required width $N$ due to an exponential increase in the number of cross-terms upon expansion; which needs to be suppressed by the factor $1/N$. We refer the reader to \Cref{app:analytical} for a detailed discussion.

\subsection{Comparison with trained networks}
\begin{figure}[!h]
    \centering
    \includegraphics[width=\textwidth]{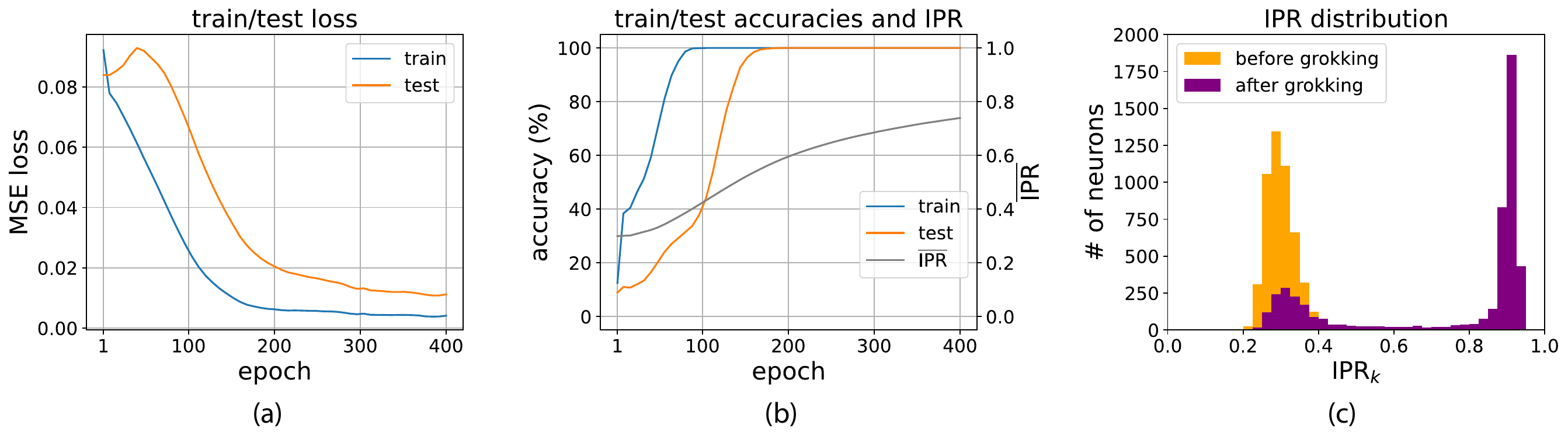}
    \caption{Training on modular addition with many terms ($(n_1+n_2+n_3+n_4)\, \mathrm{mod} \,p$). $p=11; N=5000$; Adam optimizer; learning rate $=0.005$; weight decay $=5.0$; $50\%$ of the dataset used for training. (a) MSE loss on train and test dataset. (b) Accuracy on train and test dataset as well as average IPR of the network $\overline{\mathrm{IPR}}$. The training curves show the well-known grokking phenomenon; while $\overline{\mathrm{IPR}}$ monotonically increases. (c) Initial and final IPR distributions, evidently showing periodic neurons in the grokked network, confirming the similarity to \eqref{eq:solution_multisum}. Note that the IPR for the analytical solution (\eqref{eq:solution_multisum}) is 1.}
    \label{fig:multisum}
\end{figure}

We compare the above solution to networks trained on modular addition data. We see that 2-layer MLP with sufficient width can indeed grok on modular arithmetic data. Moreover, the trained network learns qualitatively similar weights to the analytical solution \eqref{eq:solution_multisum}. The network generalizes by learning the periodic weights, as evidenced by the initial and final $\overline{\mathrm{IPR}}$ distributions in \Cref{fig:multisum}.

\section{Modular Multiplication}

Now consider the modular multiplication task in two variables\footnote{The analysis can be readily extended to modular multiplication in more than two variables; in similar vein as the modular addition analysis present in the previous section.}:
$( n_1^{a}\, n_2^{b} ) \, \mathrm{mod} \, p$ (where $a,b \in \mathbb{Z}_p \backslash \{0\}$; $p$ is a prime number). It can be learned by a 2-layer MLP, in a similar fashion as modular addition \citep{gromov2022grokking,doshi2023grok}. 
\begin{align}\label{eq:net_mul}
    \vf_{mul2} (\ve_{n_1} \oplus \ve_{n_2}) = \mQ \phi\left( \mP (\ve_{n_1} \oplus \ve_{n_2}) \right) = \mQ \phi\left( \mP^{(1)} \, \ve_{n_1} + \mP^{(2)} \, \ve_{n_2} \right)\,,
\end{align}
where we have used identical notation to \eqref{eq:net_multisum}.

\subsection{Exponential and logarithmic maps over finite fields}

Note that both modular addition and multiplication are defined over finite field of prime order $p$, denoted by $\mathcal{GF}(p)$. However, there is a subtle yet crucial difference in the two operations. Addition has a cycle of length $p$; i.e. $\left(p + p + \cdots (n \, \text{times})\right) \, \mathrm{mod} \, p = 0 \; \forall \; n \in \mathbb{Z}_p$). On the other hand, multiplication has a cycle of length $p-1$; i.e. $\left(g\cdot g \cdot \cdots (n-1 \, \text{times})\right) \; \mathrm{mod} \; p = 1$, where $g$ is a primitive root of $\mathcal{GF}(p)$.\footnote{This is a well-known result called Fermat's little theorem.} Note that multiplying any number by $0$ gives $0$; which warrants a separate treatment for $0$. The remaining elements $r \in \mathbb{Z}_p \backslash \{0\}$ can be used to define the map $g^r : \mathbb{Z}_p \backslash \{0\} \rightarrow \mathbb{Z}_p \backslash \{0\}$, where $g$ is a primitive root. This exponential is bijective (invertible). Inverting this map defines the logarithmic map with base $g$. We emphasise that over the finite field $\mathcal{GF}(p)$, the exponential ($g^r$) and logarithmic ($\log_g r$) are basically a reshuffling of the elements of $\mathbb{Z}_p \backslash \{0\}$. Reshuffling the numbers according to the logarithmic map turns the multiplication task into the addition task. This is the equivalent of the result $\log_g (n_1n_2) = \log_g n_1 + \log_g n_2$ on finite fields. Thus, barring the application of this map, the solution for multiplication is similar to \eqref{eq:solution_mul}. Note that the element $0$ needs to be treated separately while constructing the analytical solution for the network weights.

\subsection{Analytical solution}

Using the above insights, we now construct the analytical solution for the modular multiplication task. Note that in the following equation, $i \neq 0; j \neq 0; k \neq 0, q \neq 0$.

\begin{align}\label{eq:solution_mul}
\begin{aligned}
    &P^{(1)}_{00}=1\;, \qquad P^{(2)}_{00}=1\;, \qquad Q_{00}=1\;,& \\
    &P^{(1)}_{0i} = P^{(1)}_{k0} = 0 \;, \qquad P^{(2)}_{0j} = P^{(2)}_{k0} = 0 \;, \qquad Q_{q0} = Q_{k0} = 0 \;,&\\
    &\emP^{(1)}_{ki} = \left( \frac{2}{N-1} \right)^{-\frac{1}{3}} \cos{\left[ \frac{2\pi}{p-1} \sigma(k) (a\log_g{i}) + \psi^{(1)}_k \right]}\,,&\\
    &\emP^{(2)}_{kj} = \left( \frac{2}{N-1} \right)^{-\frac{1}{3}} \cos{\left[ \frac{2\pi}{p-1} \sigma(k) (b\log_g{j}) + \psi^{(2)}_k \right]}\,,&\\
    &\emQ_{qk} = \left( \frac{2}{N-1} \right)^{-\frac{1}{3}} \cos{\left[- \frac{2\pi}{p-1} \sigma(k) (\log_g{q}) - \psi^{(1)}_k - \psi^{(2)}_k \right]}\;,&
\end{aligned}
\end{align}
where $\sigma(k)$ denotes a random permutation of $k$ in $S_N$ -- reflecting the permutation symmetry of the hidden neurons. The phases $\psi_k^{(1)}$ and $\psi_k^{(2)}$ are uniformly i.i.d. sampled between $(-\pi, \pi]$.

The rows (columns) of $\mP^{(1)}, \mP^{(2)}, \mQ$ are periodic with frequencies $\frac{2\pi}{p-1}a\sigma(k), \frac{2\pi}{p-1}b\sigma(k), -\frac{2\pi}{p-1}\sigma(k)$, upon performing exponential map on their column (row) indices $i,j,q$ (Note that this excludes the $0^{th}$ entry in each row (column). Those entries deal with the element $0$). This mapping is equivalent to reshuffling the columns (rows) of $\mP^{(1)}, \mP^{(2)}, \mQ$. Let us denote these new re-shuffled, 0-excluded weight matrices as $\overline{\mP}^{(1)}, \overline{\mP}^{(2)}, \overline{\mQ}$.
\begin{align}
    &\overline{P}^{(1)}_{ki} = P^{(1)}_{k, g^i}\;,&
    &\overline{P}^{(2)}_{kj} = P^{(2)}_{k, g^j}\;,&
    &\overline{Q}_{qk} = Q_{g^q, k}&
    &(i \neq 0, j \neq 0, q \neq 0, k \neq 0) \,,&
\end{align}
where we denote the shuffled indices ($i,j,q$) by $g^i, g^j, g^q$ in the subscript.
Again, we can use IPR quantify the periodicity. We denote discrete Fourier transforms of the row- (column-) vectors $\overline{\mP}^{(1)}_{k\cdot}, \overline{\mP}^{(2)}_{k\cdot}, \overline{\mQ}_{\cdot k}$ by $\mathcal{F}\left(\overline{\mP}^{(1)}_{k\cdot}\right), \mathcal{F}\left(\overline{\mP}^{(2)}_{k\cdot}\right), \mathcal{F}\left(\overline{\mQ}_{\cdot k}\right)$.
\begin{align}\label{eq:IPR}
    &\text{IPR}\left(\overline{\mP}^{(t)}_{k\cdot}\right) \coloneqq \left(\frac{\left\lVert \mathcal{F}\left(\overline{\mP}^{(t)}_{k\cdot}\right) \right\rVert_4}{\left\lVert \mathcal{F}\left(\overline{\mP}^{(t)}_{k\cdot}\right) \right\rVert_4} \right)^2 \,;&
    &\text{IPR}\left(\overline{\mQ}_{\cdot k}\right) \coloneqq \left(\frac{\left\lVert \mathcal{F} \left(\overline{\mQ}_{\cdot k}\right) \right\rVert_4}{\left\lVert \mathcal{F} \left(\overline{\mQ}_{\cdot k}\right) \right\rVert_2} \right)^4 \,,&
\end{align}
where $t\in\{1,2\}$.
Per-neuron IPR as well as average IPR of the network can be defined as before; with $\text{IPR}_k \coloneqq 1/3 \left( \text{IPR}\left(\overline{\mP}^{(1)}_{k\cdot}\right) + \text{IPR}\left(\overline{\mP}^{(2)}_{k\cdot}\right) + \text{IPR}\left(\overline{\mQ}_{k\cdot}\right) \right)$ and $\overline{\text{IPR}} \coloneqq \mathbb E_k \left[ \text{IPR}_k \right]$.

\subsection{Comparison with trained networks}

Now, we show that the real networks trained on modular multiplication data learn similar features as \Eqref{eq:solution_mul}.

\begin{figure}[!h]
    \centering
    \includegraphics[width=\textwidth]{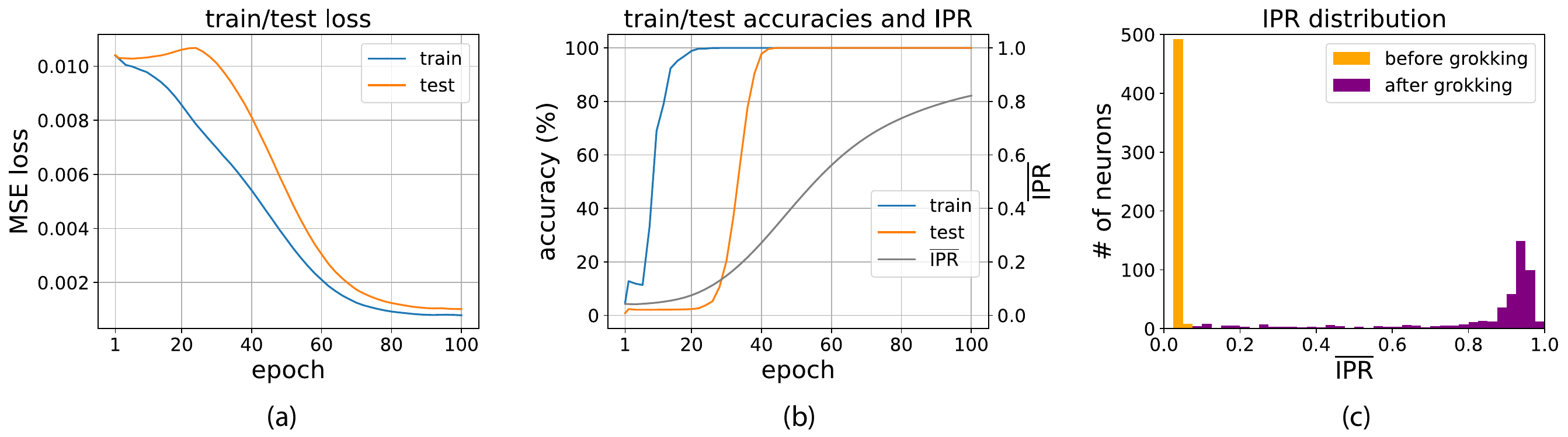}
    \caption{Training on modular multiplication ($n_1n_2\, \mathrm{mod} \,p$). $p=97; N=500$; Adam optimizer; learning rate $=0.005$; weight decay $=5.0$; $50\%$ of the dataset used for training. (a) MSE loss on train and test datset. (b) Accuracy on train and test dataset as well as average IPR of the network $\overline{\mathrm{IPR}}$. The training curves show the well-known grokking phenomenon; while $\overline{\mathrm{IPR}}$ monotonically increases. (c) Initial and final IPR distributions, evidently showing periodic neurons in the grokked network, confirming the similarity to \eqref{eq:solution_mul}. Note that the IPR for the analytical solution (\eqref{eq:solution_mul}) is 1.}
    \label{fig:multiplication}
\end{figure}

\section{Arbitrary Modular Polynomials}

Consider a general modular polynomial in two variables $(n_1,n_2)$ containing $S$ terms: 
\begin{align}
    \left( c_1 n_1^{a_1}n_2^{b_1} + c_2n_1^{a_2}n_2^{b_2} + \cdots + c_S n_1^{a_S}n_2^{b_S} \right) \, \mathrm{mod} \, p \,.
\end{align}
We utilize the solutions presented in previous sections to construct a simple network that generalizes on this task. Each term (without the coefficients $c_s$) can be solved by a 2-layer MLP expert (equations \ref{eq:net_mul}, \ref{eq:solution_mul}) that multiplies the appropriate powers of $n_1$ and $n_2$. The output of these $S$ terms can be added, along with the coefficients, using another expert network (equations \ref{eq:net_multisum}. The expert networks are designed to perform best when their inputs are close to \texttt{one\_hot} vectors; so we apply (low temperature) Softmax function to the outputs of the each term. 
As before, the input to the network are stacked, \texttt{one\_hot} represented numbers $n_1,n_2$: $(\ve_{n_1} \oplus \ve_{n_2})$
\begin{align}\label{eq:solution_arbitrary}
    &\vt^{(s)} = \vf^{(s)}_{mul2}(\ve_{n_1} \oplus \ve_{n_2}) = \mQ^{(s)} \phi\left( \mP^{(s,1)} \, \ve_{n_1} + \mP^{(s,2)} \, \ve_{n_2} \right) \,,&
    &(s \in [1,S])&\\
    & \vu^{(s)} = \texttt{softmax}_{\beta}\left(\vt^{(s)}\right)\,, & \\
    &\vz = \vf_{addS}\left( \vu^{(1)} \oplus \cdots \oplus \vu^{(S)} \right) = \mW \phi \left( \mU^{(1)} \, \vu^{(1)} + \cdots + \mU^{(S)} \, \vu^{(S)} \right)&
\end{align}
The weights $\mP^{(s,1)}, \mP^{(s,2)}, \mQ^{(s)}$ are given by \eqref{eq:solution_mul}, with appropriate powers $a_s, b_s$ for each term. Similarly, the weights $\mU^{(s)}, \mW$ are given by \eqref{eq:solution_multisum}, with coefficients $c_s$. We have used the temperature-scaled Softmax function
\begin{equation}
    \texttt{softmax}_{\beta}\left(t^{(s)}_i\right) \coloneqq \frac{e^{\beta t^{(s)}_i}}{\sum_j^p e^{\beta t^{(s)}_j}} \,.
\end{equation}
We select a high value for the inverse temperature $\beta \sim 100$ so that the intermediate outputs $\vu$ get close to \texttt{one\_hot} vectors. This results in a higher accuracy in the summation of monomials to be performed in the subsequent layers.

In \Cref{app:analytical}, we show the performance of real networks, with the solutions \eqref{eq:solution_arbitrary}, on general modular polynomials. We show that such a construction is able to learn polynomials that are un-learnable via training standard network architectures.  

Instead of using the analytical weights, one can train the ``experts": $\vf^{(s)}_{addS}, \vf_{addS}$ separately, on multiplication and addition datasets, and then combine them into a Mixture-of-Experts model.

\section{Discussion}

We have presented the analytical solutions for the weights of 2-layer MLP networks on modular multiplication using the bijective exponential and logarithmic maps on $\mathcal{GF}(p)$. We have also presented solutions for modular multiplication with many terms and arbitrary coefficients. In both cases, We have shown that real networks trained on these tasks find similar solutions. 

Using these ``expert" solutions, we have constructed a network that generalizes on arbitrary modular polynomials, given sufficient width. This network is reminiscent of Mixture-of-Expert models. Therefore, our construction opens the possibility of building such network architectures, that can learn modular arithmetic tasks that SOTA models have been unable to learn. We delegate the exploration of this avenue for future work.

On the other hand, our general formulation of analytical solutions for 2-layer MLP points to a potential classification of modular polynomials into learnable and non-learnable ones.
\begin{hyp}[Weak generalization]\label{hyp:weak}
    Consider training a 2-layer MLP network trained on a dataset consisting of modular polynomial on $\mathcal{GF}(p)$ in two variables $(n_1, n_2)$, with commonly used optimizers (SGD, Adam etc.), loss functions (MSE, CrossEntropy etc.) and regularization methods (weight decay, dropout, BatchNorm). The network achieves 100\% test accuracy if the modular polynomial is of the following form:\footnote{The networks are naturally trained on $<100\%$ of the entire dataset; and tested on the rest.}
    \begin{align}\label{eq:hyp1}
        h(g_1(n_1) + g_2(n_2)) \, \mathrm{mod} \, p \,,
    \end{align}
    where $g_1$, $g_2$ and $h$ are functions on $\mathcal{GF}(p)$.
    \begin{itemize}
        \item Note that \eqref{eq:hyp1} also includes modular multiplication: taking $g_1, g_2$ to be $\log_g(\cdot)$ and $h$ to be $exp_g(\cdot)$, where $g$ is a primitive root of $\mathcal{GF}(p)$. 
        \item If the function $h$ is invertible, then it is also possible to construct an analytical solution for this task in a similar fashion as equations \ref{eq:solution_multisum} and \ref{eq:solution_mul}.\footnote{This general construction of analytical solution was also hypothesized in \citet{gromov2022grokking}.}
    \end{itemize}
\end{hyp}

In \Cref{appsec:learnable}, we provide experimental evidence supporting this claim. We show the results of training experiments on various modular polynomials belonging to both categories. We see a clear difference in test performance on tasks that fall under the form \eqref{eq:hyp1} and those that do not.

It is possible to generalize this claim to other network architectures. General architectures such as transformers and deeper MLPs have also been unable to solve general modular arithmetic tasks that do not fall under \eqref{eq:hyp1} \citep{power2022grokking}. Moreover, \citet{doshi2023grok} hypothesized a general framework for generalization on modular arithmetic tasks in these architectures. Consequently, we conjecture that Hypothesis \ref{hyp:weak} can be extended to general architectures.
\begin{hyp}[Strong generalization]\label{hyp:strong}
    Consider training a standard neural network (MLP, Transformer etc.; not pre-trained) on a dataset consisting of modular polynomial on $\mathcal{GF}(p)$ in two variables $(n_1, n_2)$, with commonly used optimizers (SGD, Adam etc.), loss functions (MSE, CrossEntropy etc.) and regularization methods (weight decay, dropout, BatchNorm). The network achieves 100\% test accuracy if the modular polynomial is of the from of \eqref{eq:hyp1}.
\end{hyp}

While experiments validate these hypotheses, a comprehensive proof of the claim remains an open challenge. Formulating such a proof would require a robust classification of modular polynomials; as well as the understanding biases of neural network training on these tasks. We defer such an analysis to future works.

\subsubsection*{Acknowledgments}
A.G.’s work at the University of
Maryland was supported in part by NSF CAREER Award DMR-2045181, Sloan Foundation and the Laboratory for Physical Sciences through the Condensed Matter Theory Center.

\bibliography{Bibliography}
\bibliographystyle{iclr2024_conference}

\appendix

\section{Analytical solutions}\label{app:analytical}

\subsection{Modular multiplication}

Here we show that that the analytical solution presented in \eqref{eq:solution_mul} solves the modular multiplication task. Let us first calculate the network output for $n_1 \neq 0, n_2 \neq 0$. We will treat the cases with $n_1=0$ and/or $n_2=0$ separately. 

\begin{align}\label{eqapp:analytical_mul}
    \nonumber
    &\vf_{mul2} (\ve_{n_1} \oplus \ve_{n_2}) \\
    \nonumber
    &= \mQ \phi\left( \mP^{(1)} \, \ve_{n_1} + \mP^{(2)} \, \ve_{n_2} \right) \\
    \nonumber
    &= \sum_{k=1}^N Q_{qk}(P^{(1)}_{kn_1} + P^{(2)}_{kn_2})^2 \\
    \nonumber
    &= \frac{2}{N-1} \sum_{k=1}^{N-1} \cos{\left( - \frac{2\pi}{p-1}\sigma(k)\log_g q - (\psi^{(1)}_k + \psi^{(2)}_k) \right)} \cdot \\
    \nonumber
    &\qquad\qquad \cdot \left( \cos{\left( \frac{2\pi}{p-1}\sigma(k)a\log_g n_1 + \psi^{(1)}_k \right)} + \cos{\left( \frac{2\pi}{p-1}\sigma(k)b\log_g n_2 + \psi^{(2)}_k \right)} \right)^2 \\
    \nonumber
    &= \frac{2}{N-1} \sum_{k=1}^{N-1} \left\{ \frac{1}{4} \cos{\left( \frac{2\pi}{p}\sigma(k)(2a\log_g n_1 - \log_g q) + \psi^{(1)}_k - \psi^{(2)}_k  \right)} \right. \\
    \nonumber
    & \left. \qquad\qquad\quad + \frac{1}{4} \cos{\left( \frac{2\pi}{p-1}\sigma(k)(2a\log_g n_1 + \log_g q) + 3\psi^{(1)}_k + \phi^{(2)}_k \right)} \right. \\
    \nonumber
    &\left. \qquad\qquad\quad + \frac{1}{4} \cos{\left( \frac{2\pi}{p-1}\sigma(k)(2b\log_g n_2 - \log_g q) - \psi^{(1)}_k + \psi^{(2)}_k \right)} \right. \\
    \nonumber
    & \left. \qquad\qquad\quad + \frac{1}{4} \cos{\left( \frac{2\pi}{p-1}\sigma(k)(2b\log_g n_2 + \log_g q) + \psi^{(1)}_k + 3\psi^{(2)}_k \right)} \right. \\
    \nonumber
    &\left. \qquad\qquad\quad + \boxed{\frac{1}{2} \cos{\left( \frac{2\pi}{p-1}\sigma(k)(a\log_g n_1 + b\log_g n_2 - \log_g q) \right)}} \right. \\
    \nonumber
    &\left. \qquad\qquad\quad + \frac{1}{2} \cos{\left( \frac{2\pi}{p-1}\sigma(k)(a\log_g n_1 + b\log_g n_2 + \log_g q) + 2\psi^{(1)}_k + 2\psi^{(2)}_k \right)} \right. \\
    \nonumber
    &\left. \qquad\qquad\quad + \frac{1}{2} \cos{\left( \frac{2\pi}{p-1}\sigma(k)(a\log_g n_1 - b\log_g n_2 - \log_g q) - 2\psi^{(1)}_k \right)} \right.\\
    \nonumber
    &\left. \qquad\qquad\quad + \frac{1}{2} \cos{\left( \frac{2\pi}{p-1}\sigma(k)(a\log_g n_1 - b\log_g n_2 + \log_g q) + 2\psi^{(1)}_k \right)} \right. \\
    &\left. \qquad\qquad\quad + \cos{\left( - \frac{2\pi}{p-1}\sigma(k)\log_g q - \psi^{(1)}_k - \psi^{(2)}_k \right)} \right\} \,.
\end{align}

We have highlighted the term that will give us the desired output with a box. Note that the desired term is the only one that does not have additive phases in the argument. Recall that the phases $\psi^{(1)}$ and $\psi^{(2)}$ are randomly chosen -- uniformly iid sampled between $(-\pi, \pi]$. Consequently, as $N$ becomes large, all other terms will vanish due to random phase approximation. The only surviving term will be the boxed term. We can write the boxed term in a more suggestive form to make the analytical solution apparent.

\begin{align}
    \nonumber
    \vf_{mul2} (\ve_{n_1} \oplus \ve_{n_2}) 
    &= \frac{1}{N-1} \sum_{k=1}^{N-1} \cos{\left( \frac{2\pi}{p-1}\sigma(k)(a\log_g n_1 + b\log_g n_2 - \log_g q) \right)}\\
    \nonumber
    &= \frac{1}{N-1} \sum_{k=1}^{N-1} \cos{\left( \frac{2\pi}{p-1}\sigma(k)(\log_g n_1^a + \log_g n_2^b - \log_g q) \right)} \\
    &\sim \delta^p \left( n_1^a n_2^b - q \right) \,,
\end{align}
were we have defined the \emph{modular Kronecker Delta function} $\delta(\cdot)$ as Kronecker Delta function up to integer multiplies of the modular base $p$.
\begin{align}
    \delta^p(x) = 
    \begin{cases}
    1 & \; x = r p \qquad (r \in \mathbb{Z}) \\
    0 & \; \text{otherwise}
    \end{cases} \,,
\end{align}
where $\mathbb Z$ denotes the set of all integers.
Note that $\delta^p(n_1^a n_2^b - q)$ are the desired \texttt{one\_hot} encoded labels for the modular multiplication task, by definition. Thus our network output with periodic weights is indeed a solution.

Note that if either $n_1=0$ or $n_2=0$ (or both), the $0^{th}$ output logit will be the largest ($=1$). Consequently, the network will correctly predict 0 output for such cases. 

Hence, the solution presented in \eqref{eq:solution_mul} gives $100\%$ accuracy .

\subsection{Modular addition with many terms}

Next, we show that that the analytical solution presented in \eqref{eq:solution_multisum} solves the task of modular addition with many terms.

\begin{align}\label{eqapp:analytical_mul}
    \nonumber
    &\vf_{addS} (\ve_{n_1} \oplus \cdots \oplus \ve_{n_S}) \\
    \nonumber
    &= \mW \phi\left( \mU^{(1)} \, \ve_{n_1} + \cdots + \mU^{(S)} \, \ve_{n_S} \right) \\
    \nonumber
    &= \sum_{k=1}^N W_{qk}(U^{(1)}_{kn_1} + \cdots  P^{(S)}_{kn_S})^S \\
    \nonumber
    &= \frac{2^S}{N\cdot S!}\sum_{k=1}^{N} \cos{\left( - \frac{2\pi}{p}\sigma(k)q - \sum_{s=1}^S\psi^{(s)}_k \right)} \cdot \\
    \nonumber
    &\qquad\qquad \cdot \left( \cos{\left( \frac{2\pi}{p}\sigma(k) c_1 n_1 + \psi^{(1)}_k \right)} + \cdots + \cos{\left( \frac{2\pi}{p}\sigma(k) c_2 n_2 + \psi^{(S)}_k \right)} \right)^S \\
    &= \frac{2^S}{N\cdot S!}
    \sum_{k=1}^{N} \left\{ \cdots + \boxed{\frac{S!}{2^S} \cos{\left( \frac{2\pi}{p}\sigma(k)(c_1 n_1 + \cdots + c_S n_S - q) \right)}} + \cdots \right\}
\end{align}
Here, we have omitted all the additional terms that drop out due to random phase approximation; showing only the desired surviving term. Note that the number of these additional terms increases exponentially in $S$; and they are suppressed by a factor of $1/N$. This provides an explanation for an exponential behaviour in the required network width $N$ with increasing number of terms in the addition task \Cref{fig:multisum_analytical}.

\begin{align}
    \nonumber
    \vf_{addS} (\ve_{n_1} \oplus \cdots \oplus \ve_{n_S}) 
    &= \frac{1}{N} \sum_{k=1}^N \cos{\left( \frac{2\pi}{p}\sigma(k)(c_1 n_1 + \cdots + c_S n_S - q) \right)}\\
    \nonumber
    &\sim \delta^p \left( c_1n_1 + \cdots + c_Sn_S - q \right) \,.
\end{align}

Note that $\delta^p \left( c_1n_1 + \cdots + c_Sn_S - q \right)$ is indeed the desired output for the modular addition task.

\section{Performance on arbitrary modular polynomials}
\label{appsec:arbitrary}

In this appendix, we document the performance of the network constructed in \eqref{eq:solution_arbitrary} on general modular polynomials. Note that none of the polynomials listed in \Cref{tab:arbitrary} are learnable by training 2-layer MLP or other network architectures (such as Transformers). The MSE loss and accuracy are computed on the entire dataset ($p^2$ examples). The expert-widths are taken to be $N_1=500$ and $N_2=2000$. Since $S=3$ in all the examples, we use cubic ($\phi(x) = x^3$) activation function. We have truncated the MSE loss at 6 digits after the decimal point.

\begin{center}\label{tab:arbitrary}
\begin{tabular}{|c||c|c|}
    \hline
     Modular polynomial & MSE loss & Accuracy  \\
     \hline
     $\left(2n_1^4 n_2 + n_1^2 n_2^2 + 3n_1 n_2^3\right) \, mod \, 97$ & 0.007674 & $100\%$ \\
     $\left(n_1^5 n_2^3 + 4n_1^2 n_2 + 5n_1^2 n_2^3\right) \, mod \, 97$ & 0.007660 & $100\%$ \\
     $\left(7n_1^4 n_2^4 + 2n_1^3 n_2^2 + 4n_1^2 n_2^5\right) \, mod \, 97$ & 0.007683 & $100\%$ \\
     $\left(2n_1^4 n_2 + n_1^2 n_2^2 + 3n_1 n_2^3\right) \, mod \, 23$ & 0.009758 & $100\%$ \\
     $\left(n_1^5 n_2^3 + 4n_1^2 n_2 + 5n_1^2 n_2^3\right) \, mod \, 23$ & 0.009757 & $100\%$ \\
     $\left(7n_1^4 n_2^4 + 2n_1^3 n_2^2 + 4n_1^2 n_2^5\right) \, mod \, 23$ & 0.010201 & $100\%$ \\
     \hline
\end{tabular}
\end{center}

\section{Training on learnable and non-learnable modular polynomials}
\label{appsec:learnable}

Here we present the training results on various modular polynomials. 2-layer MLP networks with quadratic activations and width $N=5000$ are trained Adam optimizer, MSE loss, learning rate $=0.005$, weight decay $=5.0$, on $50\%$ of the total dataset.

We observe that training the network on polynomials of the form $h(g_1(n_1) + g_2(n_2)) \, \mathrm{mod} \, p$ results in generalization. Whereas, making a slight change to these polynomials results in inability of the networks to generalize. This serves as evidence for Hypothesis \ref{hyp:weak}.

Note that the following results remain qualitatively unchanged upon changing/tuning hyperparameters.

% \begin{tabular}{|c||c|c|c|c|}
%     \hline
%     Modular polynomial & Train loss & Test loss & Train acc & Test acc \\
%     \hline
%     $\left(4n_1 + n_2^2\right)^3 \, \mod \, 97$ & 0.00034401542507112026 & 0.0005693577113561332 & $100\%$ & $100\%$ \\
%     \textcolor{red}{$\left(4n_1 + n_2^2\right)^3 + n_1n_2 \, \mod \, 97$} & \textcolor{red}{0.001963343471288681} & \textcolor{red}{0.011216854676604271} & \textcolor{red}{$100\%$} & \textcolor{red}{$2.27\%$} \\
%     $\left(2n_1 + 3n_2\right)^4 \, \mod \, 97$ & 0.00013941623910795897 & 0.00017259560991078615 & $100\%$ & $100\%$ \\
%     \textcolor{red}{$\left(2n_1 + 3n_2\right)^4 - n_1^2 \, \mod \, 97$} & \textcolor{red}{0.0019169057486578822} & \textcolor{red}{0.011097615584731102} & \textcolor{red}{$100\%$} & \textcolor{red}{$3.93\%$} \\
%     $\left(4n_1 + n_2^2\right)^3 \, \mod \, 23$ & 0.0001430085249012336 & 0.006274334155023098 & $100\%$ & $100\%$ \\
%     \textcolor{red}{$\left(4n_1 + n_2^2\right)^3 + n_1n_2 \, \mod \, 97$} & \textcolor{red}{0.0001478071790188551} & \textcolor{red}{0.05186643451452255} & \textcolor{red}{$100\%$} & \textcolor{red}{$1.89\%$} \\
%     $\left(2n_1 + 3n_2\right)^4 \, \mod \, 23$ & 9.375475201522931e-05 & 0.001010152860544622 & $100\%$ & $100\%$ \\
%     \textcolor{red}{$\left(2n_1 + 3n_2\right)^4 - n_1^2 \, \mod \, 23$} & \textcolor{red}{0.00013254278746899217} & \textcolor{red}{0.049618810415267944} & \textcolor{red}{$100\%$} & \textcolor{red}{$7.17\%$} \\
%     \hline
% \end{tabular}

\begin{tabular}{|c||c|c|c|c|}
    \hline
    Modular polynomial & Train loss & Test loss & Train acc & Test acc \\
    \hline
    $\left(4n_1 + n_2^2\right)^3 \, \mod \, 97$ & 0.000344 & 0.000569 & $100\%$ & $100\%$ \\
    \textcolor{red}{$\left(4n_1 + n_2^2\right)^3 + n_1n_2 \, \mod \, 97$} & \textcolor{red}{0.001963} & \textcolor{red}{0.011216} & \textcolor{red}{$100\%$} & \textcolor{red}{$2.27\%$} \\
    $\left(2n_1 + 3n_2\right)^4 \, \mod \, 97$ & 0.000139 & 0.000172 & $100\%$ & $100\%$ \\
    \textcolor{red}{$\left(2n_1 + 3n_2\right)^4 - n_1^2 \, \mod \, 97$} & \textcolor{red}{0.001916} & \textcolor{red}{0.011097} & \textcolor{red}{$100\%$} & \textcolor{red}{$3.93\%$} \\
    $\left(5n_1^3 + 2n_2^4\right)^2 \, \mod \, 97$ & 0.000146 & 0.000161 & $100\%$ & $100\%$ \\
    \textcolor{red}{$\left(5n_1^3 + 2n_2^4\right)^2 - n_2 \, \mod \, 97$} & \textcolor{red}{0.001523} & \textcolor{red}{0.006108} & \textcolor{red}{$100\%$} & \textcolor{red}{$72.32\%$} \\
    $\left(4n_1 + n_2^2\right)^3 \, \mod \, 23$ & 0.000143 & 0.006274 & $100\%$ & $100\%$ \\
    \textcolor{red}{$\left(4n_1 + n_2^2\right)^3 + n_1n_2 \, \mod \, 23$} & \textcolor{red}{0.000147} & \textcolor{red}{0.051866} & \textcolor{red}{$100\%$} & \textcolor{red}{$1.89\%$} \\
    $\left(2n_1 + 3n_2\right)^4 \, \mod \, 23$ & 0.000093 & 0.001010 & $100\%$ & $100\%$ \\
    \textcolor{red}{$\left(2n_1 + 3n_2\right)^4 - n_1^2 \, \mod \, 23$} & \textcolor{red}{0.000132} & \textcolor{red}{0.049618} & \textcolor{red}{$100\%$} & \textcolor{red}{$7.17\%$} \\
    $\left(5n_1^3 + 2n_2^4\right)^2 \, \mod \, 23$ & 0.000056 & 0.001004 & $100\%$ & $100\%$ \\
    \textcolor{red}{$\left(5n_1^3 + 2n_2^4\right)^2 - n_2 \, \mod \, 23$} & \textcolor{red}{0.000150} & \textcolor{red}{0.052047} & \textcolor{red}{$100\%$} & \textcolor{red}{$2.64\%$} \\
    \hline
\end{tabular}

\end{document}